%% file: ACM-MM-main.tex
\documentclass[sigconf]{acmart}

\AtBeginDocument{%
  }

\usepackage{multirow}
\usepackage{subfigure}

\usepackage{amssymb}
\usepackage{amsmath}
\usepackage{listings}
\usepackage[linesnumbered,ruled,vlined]{algorithm2e}

\usepackage{tcolorbox}
\usepackage[utf8]{inputenc}

\usepackage{hyperref}
\usepackage{fontawesome}

\setcopyright{acmlicensed}
\copyrightyear{2025}
\acmYear{2025}
\acmDOI{XXXXXXX.XXXXXXX}
\acmConference[ACM MM '25 Dataset Track]{ACM International Conference on Multimedia (Dataset Track)}{October 2025}{Lisbon, Portugal}
\acmISBN{978-1-4503-XXXX-X/2018/06}

\begin{document}

\title{Evaluating MLLMs with \\ \underline{M}ultimodal \underline{M}ulti-image \underline{R}easoning \underline{B}enchmark}

\author{
Ziming Cheng\textsuperscript{1,\dag,\S}, Binrui Xu\textsuperscript{1,\S}, Lisheng Gong\textsuperscript{1,\S}, Zuhe Song\textsuperscript{1,\S}, 
Tianshuo Zhou\textsuperscript{1,*}, Shiqi Zhong\textsuperscript{1,*}, Siyu Ren\textsuperscript{1,*}, Mingxiang Chen\textsuperscript{1,*}, 
Xiangchao Meng\textsuperscript{1,*}, Yuxin Zhang\textsuperscript{2,*}, Yanlin Li\textsuperscript{3,*}, 
Lei Ren\textsuperscript{4}, Wei Chen\textsuperscript{4}, 
Zhiyuan Huang\textsuperscript{5}, Mingjie Zhan\textsuperscript{5}, 
Xiaojie Wang\textsuperscript{1}, Fangxiang Feng\textsuperscript{1}
}

\affiliation{
\textsuperscript{1}BUPT \country{China} \hspace{0.5em}
\textsuperscript{2}YSU \country{China} \hspace{0.5em}
\textsuperscript{3}NUS \country{Singapore} \hspace{0.5em}
\textsuperscript{4}Li Auto Inc. \country{China} \hspace{0.5em}
\textsuperscript{5}SenseTime Research \country{China}\\
\textsuperscript{\S}Core contribution \quad
\textsuperscript{*}Equal contribution \quad
\textsuperscript{\dag}Project lead  \quad \href{https://github.com/LesterGong/MMRB}{\faGithub\ https://github.com/LesterGong/MMRB}
}

\renewcommand{\shortauthors}{Ziming Cheng et al.}


\input{latex/00-Abstract}



\received{31 May 2025}
\received[revised]{12 March 2009}
\received[accepted]{5 June 2009}

\maketitle
\input{latex/figures/main-figure}
\input{latex/tables/mmrb-compare}
\input{latex/01-Introduction}
\input{latex/02-Related-Work}
\input{latex/03-Pilot-Experiment}
\input{latex/figures/data-pipeline}
\input{latex/04-Multimodal-Multi-image-Reasoning-Benchmark}
\input{latex/05-Evaluation-Metrics}

\input{latex/06-Experimental-Setup}
\input{latex/tables/reward-model-results}
\input{latex/07-Main-Results-and-Discussion}

\input{latex/figures/model-size-vs-efficacy}
\input{latex/08-Conclusion}


\clearpage
\bibliographystyle{ACM-Reference-Format}
\bibliography{acmart}

\clearpage
\appendix
\input{latex/Appendix/_Appendix-Overview}
\input{latex/Appendix/A-Minimum-Sample-Capacity-Estimation}
\input{latex/Appendix/B-Evaluation-Metrics}
\input{latex/Appendix/C-Implementation-Details}
\input{latex/Appendix/D-Error-Analysis}
\input{latex/Appendix/E-Data-Samples}
\input{latex/Appendix/F-Data-Sources}
\input{latex/Appendix/G-Baseline-Models}
\input{latex/Appendix/H-Prompt-Templates}

\end{document}

%% file: latex/00-Abstract.tex
\begin{abstract}

With enhanced capabilities and widespread applications, Multimodal Large Language Models (MLLMs) are increasingly required to process and reason over multiple images simultaneously. However, existing MLLM benchmarks focus either on single-image visual reasoning or on multi-image understanding tasks with only final-answer evaluation, leaving the reasoning capabilities of MLLMs over multi-image inputs largely underexplored. To address this gap, we introduce the \textbf{Multimodal Multi-image Reasoning Benchmark (MMRB)}, the first benchmark designed to evaluate structured visual reasoning across multiple images. MMRB comprises \textbf{92 sub-tasks} covering spatial, temporal, and semantic reasoning, with multi-solution, CoT-style annotations generated by GPT-4o and refined by human experts. A derivative subset is designed to evaluate multimodal reward models in multi-image scenarios. To support fast and scalable evaluation, we propose a sentence-level matching framework using open-source LLMs. Extensive baseline experiments on \textbf{40 MLLMs}, including 9 reasoning-specific models and 8 reward models, demonstrate that open-source MLLMs still lag significantly behind commercial MLLMs in multi-image reasoning tasks. Furthermore, current multimodal reward models are nearly incapable of handling multi-image reward ranking tasks. 

\end{abstract}

%% file: latex/figures/main-figure.tex
\begin{figure}[t]
  \centering
  \includegraphics[width=\linewidth]{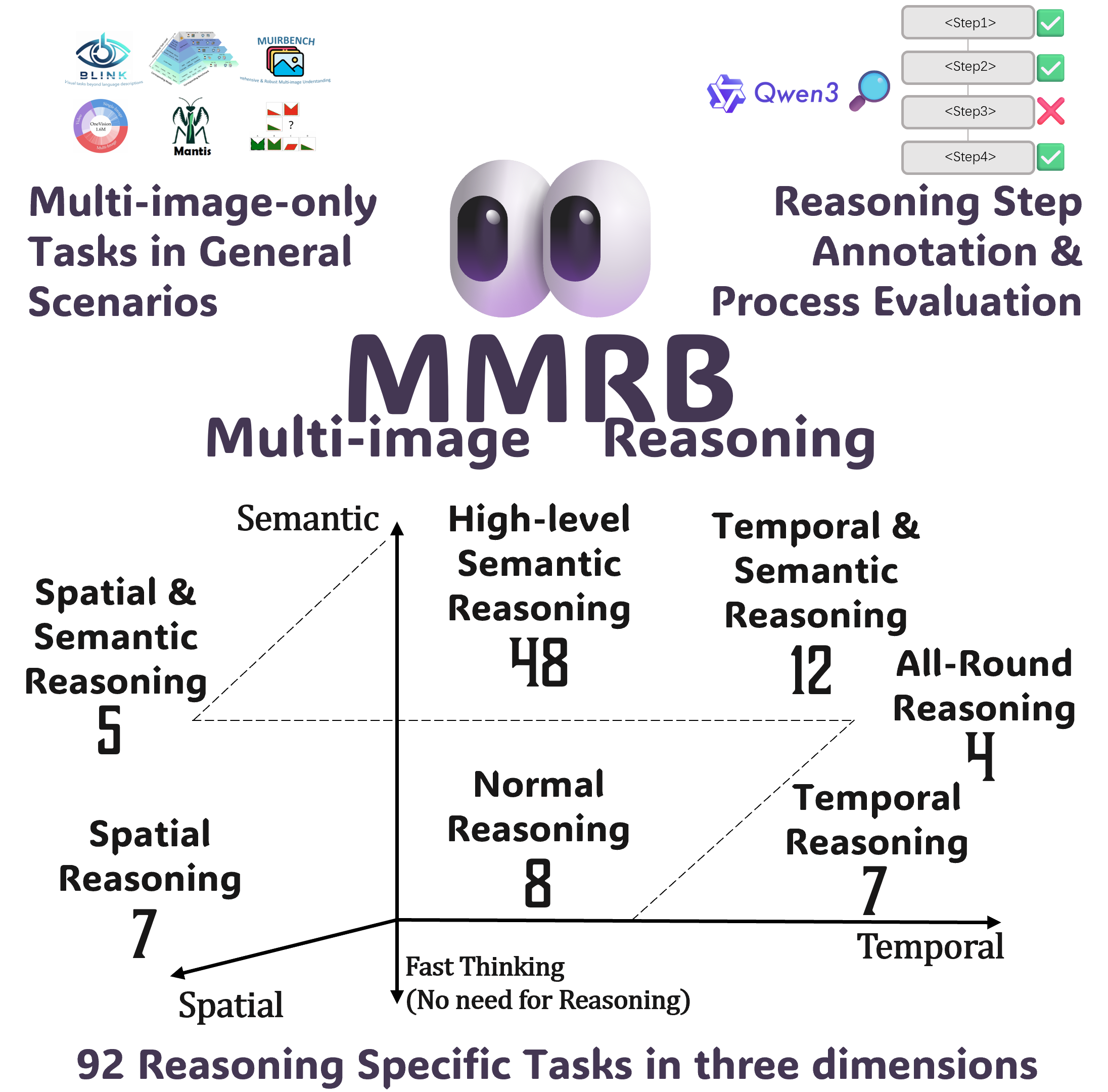}
  \caption{Overview of the MMRB benchmark, which evaluates MLLMs on 92 multi-image-only sub-tasks annotated with reasoning steps.}
  \label{fig:main-figure}
\end{figure}

%% file: latex/tables/mmrb-compare.tex
\begin{table*}[t]
\begin{center}
\small
\begin{tabular}{ccccccccc}
\hline
\textbf{Datasets}  & \textbf{Year} & \textbf{Domain}        & \textbf{\begin{tabular}[c]{@{}c@{}}Num\\ Sub-tasks\end{tabular}} & \textbf{\begin{tabular}[c]{@{}c@{}}Num\\ Samples\end{tabular}} & \textbf{\begin{tabular}[c]{@{}c@{}}Total\\ Reasoning Steps\end{tabular}} & \textbf{\begin{tabular}[c]{@{}c@{}}Avg\\ Reasoning Steps\end{tabular}} & \textbf{\begin{tabular}[c]{@{}c@{}}Avg\\ Images\end{tabular}} & \textbf{\begin{tabular}[c]{@{}c@{}}Multiple\\ Solutions\end{tabular}} \\ \hline
CoMT\cite{cheng2025comt}             & 2024          & Math, General          & 4                                                                & 3,853                                                          & 29,706                                                                   & 7.71                                                                   & 3.84                                                          & $\times$                                                                     \\
WorldQA\cite{zhang2024worldqa}            & 2024          & General                & 1                                                                & 1,007                                                          & 4,481                                                                    & 4.45                                                                   & 2.41                                                          & $\times$                                                                     \\
MiCEval\cite{zhou2024miceval}            & 2024          & General                & 8                                                                & 643                                                            & 2,889                                                                    & 4.49                                                                   & 1                                                             & $\times$                                                                     \\
MME-CoT\cite{jiang2025mme-cot}            & 2025          & Science, Math, General & 23                                                               & 1,130                                                          & 3,865                                                                    & 3.42                                                                   & 2.95                                                          & $\times$                                                                     \\
VisualProcessBench\cite{wang2025visualprm} & 2025          & Math, Science          & 5                                                                & 2,866                                                          & 26,950                                                                   & 9.40                                                                   & 1                                                             & $\times$                                                                     \\
SVIP-Test\cite{gao2025svip}          & 2025          & General                & 10                                                               & 1,934                                                          & 5,509                                                                    & 2.85                                                                   & 1                                                             & $\times$                                                                     \\
M3CoT\cite{chen2024m3cot}              & 2025          & Science, Math, General & 17                                                               & 11,459                                                         & 124,903                                                                  & 10.90                                                                  & 0.98                                                          & $\times$                                                                     \\
VRC-Bench\cite{thawakar2025llamav-o1}          & 2025          & Science, Math, General & 8                                                                & 1000                                                           & 4173                                                                     & 4.17                                                                   & 1                                                             & $\times$                                                                     \\ \hline
MMRB (Ours)        & 2025          & General                & 92                                                               & 4750                                                           & 68,882                                                                   & 4.83                                                                   & 6.17                                                          & Avg 1.93                                                              \\ \hline
\end{tabular}\end{center}
\caption{Comparison of MMRB with recently published visual reasoning benchmarks.}
\label{mmrb-compare}
\end{table*}

%% file: latex/01-Introduction.tex
\section{Introduction}
\label{intro}

Multimodal Large Language Models (MLLMs), represented by OpenAI o1\cite{jaech2024openai-o1} and Gemini 2.5 Pro \cite{anil2023gemini}, exhibit extraordinary performance in visual reasoning tasks and have fueled a boom in research on open-source visual reasoning models based on Reinforcement Learning (RL). By incorporating Chain-of-Thought (CoT) \cite{wei2022cot} or Multimodal-CoT (MCoT) \cite{zhang2023mcot} techniques, models such as LLaVA-o1 \cite{xu2024llava-o1} and LlamaV-o1 \cite{thawakar2025llamav-o1}, which adopt a systematic and structured multimodal reasoning process, demonstrate promising potential in approaching the capabilities of OpenAI o1.

To thoroughly and systematically evaluate the performance of these MLLMs, a number of benchmarks involving image contexts have been published in parallel. These evaluative efforts fall into two main directions. One focuses on complex visual reasoning problems such as science-related question answering (e.g., ScienceQA \cite{saikh2022scienceqa}, MMMU \cite{yue2024mmmu}) and mathematical reasoning, particularly in algebra (e.g., MathVista \cite{lu2023mathvista}, MathVision \cite{wang2024mathvision}), featuring extensive annotations of various intermediate reasoning steps and evaluation protocols supported by LLMs. Their evaluation domains have recently been extended to more general scenarios by works such as M3CoT \cite{chen2024m3cot}, MME-CoT \cite{jiang2025mme-cot}, and CoMT \cite{cheng2025comt}, although they still primarily focus on a single-image setting. Another direction expands the domain by introducing multi-image understanding tasks. Works such as BLINK \cite{fu2024blink} and MuirBench \cite{wang2024muirbench} collected multi-image tasks from publicly available sources, MIBench \cite{liu2024mibench} introduced image-attached In-Context Learning (ICL) tasks, and MMIU \cite{meng2024mmiu} further aggregated such tasks from existing benchmarks, covering a total of 52 tasks.

However, all existing MLLM benchmarks in the literature involve either single-image visual reasoning tasks or multi-image understanding tasks that include only answer annotations without reasoning steps, largely leaving the evaluation of visual reasoning in multi-image scenarios underexplored and making the reasoning quality of MLLMs with multi-image inputs an open question. 

Motivated by this observation, we introduce a novel \textbf{Multimodal Multi-image Reasoning Benchmark (MMRB)}, which is the first to combine visual reasoning and multi-image understanding tasks to evaluate MLLMs’ reasoning capability. Specifically, we collect a total of \textbf{92 reasoning-specific sub-tasks} from prior multi-image datasets and benchmarks that cover spatial, temporal, and high-level semantic reasoning dimensions \citep{meng2024mmiu}. We then annotate the intermediate reasoning paths using GPT-4o and refine them through expert human corrections. All reasoning paths involved follow structured and systematic MCoT steps, and additionally retain multiple possible thinking and solutions. As shown in Table \ref{mmrb-compare}, MMRB stands out as the largest benchmark by sub-task count and image density, the only one to offer multiple-solution annotations. Using reasoning steps initially annotated by GPT-4o and subsequently corrected by humans, we further construct a subset for evaluating the performance of multimodal reward models in multi-image contexts—the first of its kind in the literature.

To comprehensively evaluate MLLMs' performance on MMRB and democratize the evaluation process, we developed an \textbf{open-sourced LLM-based evaluator} that matches CoT answers with annotated ground-truth reasoning steps at the sentence level. This approach avoids expensive and time-consuming GPT-based evaluation while retaining high evaluation quality.
We evaluated a diverse set of \textbf{40 commercial and open-source MLLMs} with support for multi-image input, including 9 reasoning-specific models and 8 multimodal reward models on our MMRB benchmark, yielding several key findings: (1) Model, data, and inference-time scaling all improve performance on multi-image reasoning tasks. (2) CoT prompting boosts accuracy more substantially for non-reasoning models, while still being beneficial for reasoning-specific models. (3) Multimodal reward models exhibit instability in multi-image reward tasks.

%% file: latex/02-Related-Work.tex
\section{Related Work}
\label{related-work}

\subsection{Visual Reasoning Benchmarks}

Visual reasoning is the cognitive process to solve multi-modal problems through processing and manipulating visual information, and is the core capability of both human and advanced MLLMs \cite{xu2025visulogic}. In recent years, the evaluation of MLLMs' performance on visual reasoning tasks has gained attention. This line of research initially focused on mathematical problem-solving, such as algebra \cite{wang2024mathvision}, and has since expanded to the domains including scientific reasoning \cite{saikh2022scienceqa}, abstract logic \cite{kazemi2024remi}, and general scene understanding  \cite{zhang2024worldqa} (See Table \ref{mmrb-compare}). The evaluation objectives have progressed from an exclusive focus on final answer accuracy to also incorporating metrics that assess the correctness of intermediate reasoning steps. MME-CoT \cite{jiang2025mme-cot} further extends this by evaluating the quality, robustness, and efficiency of reasoning processes for a more comprehensive and fine-grained assessment.

However, most existing benchmarks primarily evaluate the visual reasoning capabilities of MLLMs in single-image scenarios, while overlooking tasks that involve multiple images as contexts. As a result, they offer limited insight into MLLMs’ general visual reasoning abilities under the emerging "One-Vision" framework \cite{li2024llava-onevision, Qwen2.5-VL, zhu2025internvl3}. Additionally, these benchmarks often contain relatively smaller and domain-specific task sets, potentially restricting their generalizability.

Table \ref{mmrb-compare} quantitatively compares the key attributes of prior benchmarks and our work. A few benchmarks that include math or video tasks—such as CoMT \cite{cheng2025comt}, WorldQA \cite{zhang2024worldqa}, and MME-CoT \cite{jiang2025mme-cot}—contain a small number of multi-image tasks. All other benchmarks feature only one or fewer images per question and include at most 23 sub-tasks. In contrast, our MMRB benchmark is meticulously curated with a dedicated set of multi-image tasks, addressing this gap through 92 reasoning-specific evaluation tasks.

\subsection{Multi-image Understanding Benchmarks}

Multi-image understanding tasks challenge models to integrate and analyze contextual cues from a variety of semantic, spatial, and temporal perspectives \cite{wang2024muirbench}. Research on benchmarks of this kind has seen substantial development over the past few years, ranging from visual difference spotting \cite{suhr2019nlvr2, goldstein2009birdstwords}, to video understanding \cite{wang2024mementos, li2024mvbench, song2024milebench, ding2023mevis}, and observation of low-level features \cite{wu2023qbench, fu2024blink}. Large-scale benchmarks such as MMIU \cite{meng2024mmiu} have summed up previous tasks and provide a comprehensive evaluation with 52 sub-tasks.

Multi-image in-context learning is another key capability introduced by Flamingo \cite{alayrac2022flamingo} and evaluated by benchmarks such as MIBench \cite{liu2024mibench}. Instruction-following datasets and benchmarks such as DEMON \cite{li2023DEMON} and MANTIS \cite{jiang2024mantis} demonstrate transfer learning from single-image models to perform multi-image tasks. LLaVA-OneVision \cite{li2024llava-onevision} further proposed a framework that incorporates single, multi-image, and video inputs into the training dataset and evaluates generalization across these task settings.

However, all previous multi-image benchmarks (except those limited to the domain of math \cite{lu2023mathvista, wang2024mathvision}) do not evaluate reasoning steps that involve grounding information from specific images and deriving final answers through step-by-step reasoning. We propose MMRB with the evaluation of intermediate reasoning steps to fill this gap between multi-image understanding and visual reasoning benchmarks, enabling a comprehensive assessment of how MLLMs perform reasoning in multi-image tasks.

\subsection{Multimodal Reward Model and Benchmarks}

In recent years, Reinforcement Learning from Human Feedback (RLHF) has been increasingly used to align MLLMs with human preferences \cite{schulman2017PPO, rafailov2023DPO, wang2024MPO}. Building on already powerful instruction-tuned models, this approach aims to further improve the performance upper bounds of MLLMs \cite{chen2024internvl2_5, zhu2025internvl3, sun2024LLava-RLHF}. Critic models estimate the quality of the outcome \cite{friedman2018Multi-Objective-Reward, zhang2024generative-verifiers} or the reasoning process \cite{xiong2024llava-critic, wang2025visualprm, zang2025IXC25-Reward, zhang2025r1-Reward}. Works like LLaVA-Critic \cite{xiong2024llava-critic}, VisualPRM400K \cite{wang2025visualprm}, and VLFeedback \cite{li2024vlfeedback} construct large-scale instruction-following datasets for critic training using AI-assisted methods. Meanwhile, high-quality benchmarks such as VisualPRMBench \cite{wang2025visualprm} and Multimodal RewardBench \cite{yasunaga2025multimodal-rewardbench} have been developed by human experts to evaluate critic models.

Our MMRB benchmark includes a subset constructed from pairs of incorrect AI-generated answers and their carefully corrected versions by human annotators, aiming to establish a novel multi-modal reward benchmark specifically for multi-image scenarios that have not been considered in prior research.

%% file: latex/03-Pilot-Experiment.tex
\section{Pilot Experiment}
\label{pilot-experiment}

Comprehensiveness and evaluation efficiency are crucial for a benchmark. Therefore, we first address two key design issues before constructing the full benchmark: (1) the choice of reasoning specific evaluation sub-tasks, and (2) the sample capacity for each sub-task.

\subsection{Reasoning Specific Sub-Task Selection}

\input{latex/figures/task-type}

Based on a comprehensive literature review of prior datasets and benchmarks on multi-image understanding tasks (see Appendix \ref{appx:data-sources}), we identified and collected 242 unique task candidates involving multi-image attributes without redundancy. To determine their suitability for evaluating visual reasoning, we categorized them into perceptually driven tasks and/or semantically driven tasks. Perceptually driven tasks require the model to interpret low-level visual features, whereas semantically driven tasks involve reasoning over abstract concepts, as illustrated in Figure \ref{fig:task-type.png}. In the subsequent dataset construction process, we examined the task descriptions and samples to retain only the semantically driven tasks as our reasoning-specific sub-tasks.

\subsection{Sample Capacity Estimation}

To optimize the cost efficiency of our benchmark, we estimate the minimum required sample size for each sub-task through another pilot experiment. Based on a 95\% confidence interval, we find that a minimum of 18 samples is needed for the outcome score and 10 for the process score to achieve stable evaluation results. To ensure robustness, we ultimately set a uniform sample size of 50, which lies well within the safe zone. 

For details and formulas, please see Appendix \ref{appx:minimum-sample-capacity}.

%% file: latex/figures/task-type.tex
\begin{figure}[h]
  \centering
  \includegraphics[width=\linewidth]{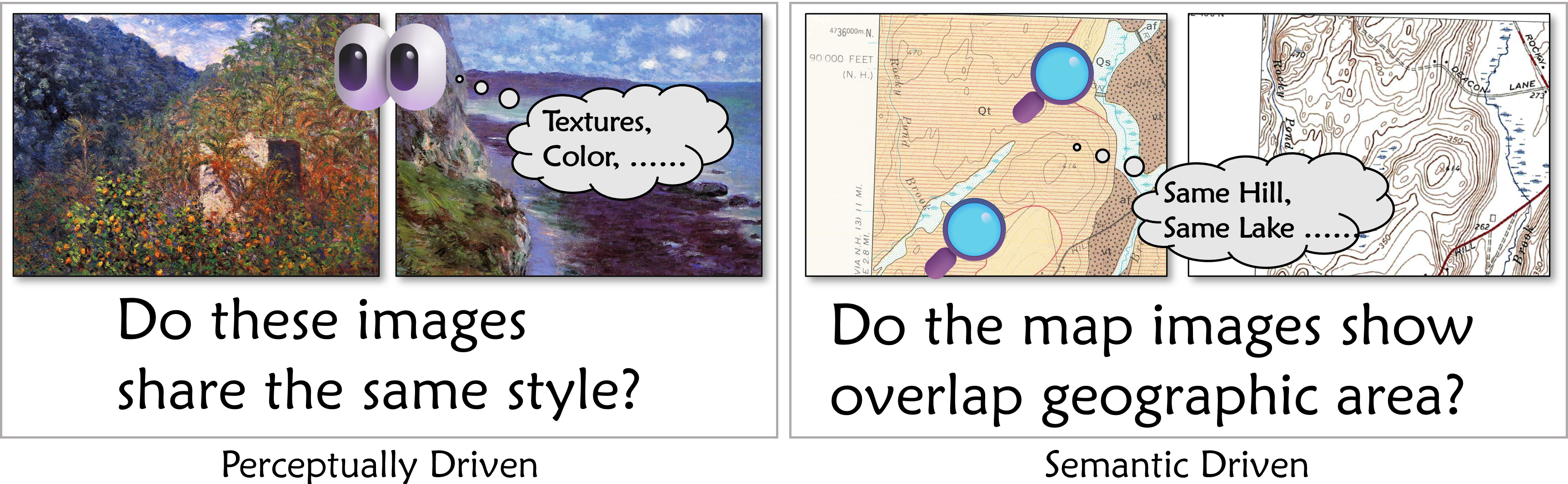}
  \caption{Perceptually driven VS. Semantically driven tasks.}
  \label{fig:task-type.png}
\end{figure}

%% file: latex/figures/data-pipeline.tex
\begin{figure*}[t]
  \centering
  \includegraphics[width=\linewidth]{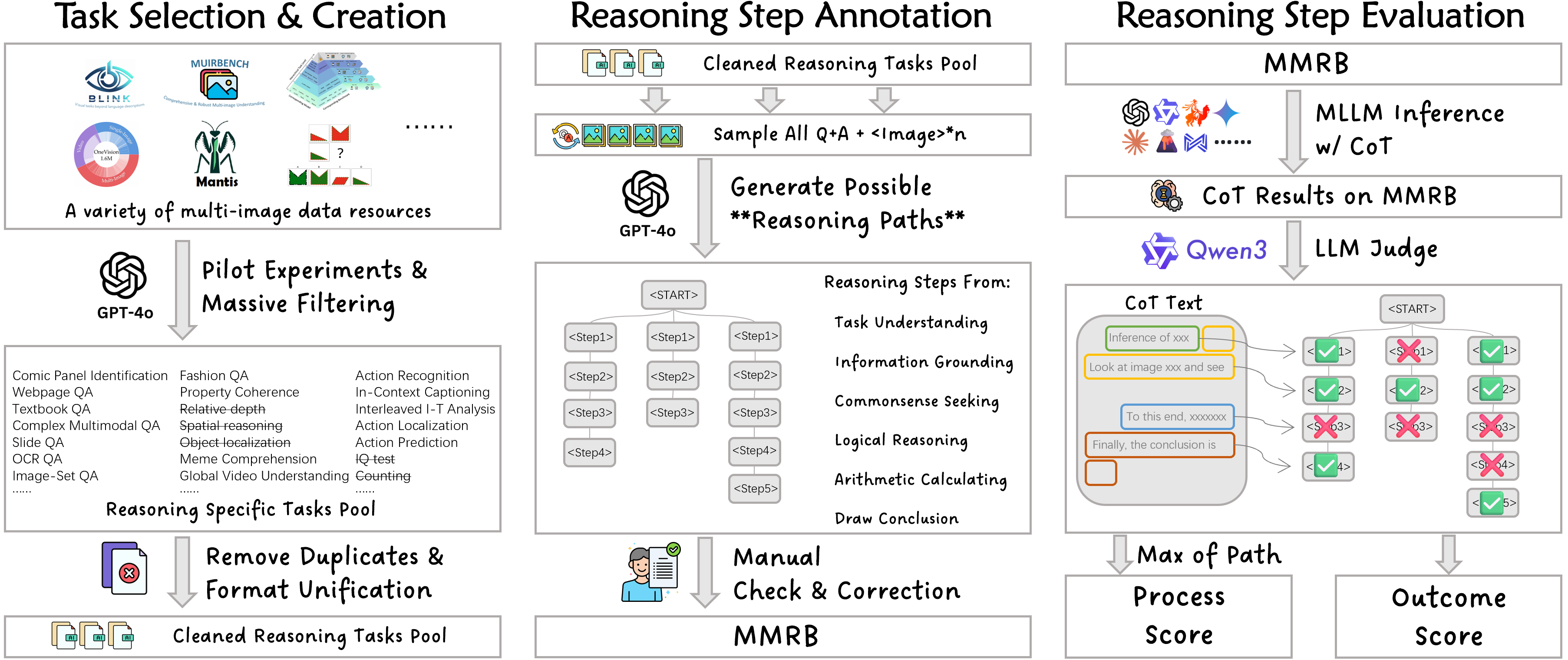}
  \caption{Data annotation pipeline of MMRB benchmark and the evaluation process.}
  \label{fig:data-pipeline}
\end{figure*}

%% file: latex/04-Multimodal-Multi-image-Reasoning-Benchmark.tex
\section{\underline{M}ultimodal \underline{M}ulti-image \underline{R}easoning \underline{B}enchmark}

\label{dataset-constrction}

\subsection{Overview of MMRB}

We first present an overview of our Multimodal Multi-image Reasoning Benchmark. The key statistics are shown in Table \ref{dataset-statistics}. Our benchmark consists of 4,750 samples encompassing 68,882 reasoning steps across 92 sub-tasks, covering semantic, spatial, and temporal reasoning. Notably, each sample contains an average of 6.17 images and 1.93 distinct solutions. During the annotation process, we also corrected a horrifying 355 incorrect ground truths from the source datasets, which could have resulted in up to a 14\% deviation in our benchmark if left uncorrected.

\input{latex/tables/dataset-statistics}

\subsection{Data Construction Pipeline}

We illustrate the data construction pipeline of MMRB in this section, which consists three main stages: task selection \& creation, reasoning step annotation, and manual correction, as the left two blocks of Figure \ref{fig:data-pipeline} shows.

\subsubsection{\textbf{Sub-task Selection and Data Source Preparation}}

Based on a comprehensive literature review of previous datasets and benchmarks on multi-image understanding tasks (catalogs see Appendix \ref{appx:data-sources}), we identified and collected 242 unique task candidates from 22 datasets, all involving multi-image contexts and selected without redundancy. After ensuring license compliance and access availability, these datasets were extensively downloaded, reduplicated, and categorized into high-level semantic, temporal, or spatial reasoning types, as the taxonomy proposed by MMIU \cite{meng2024mmiu}.

Following the sub-task selection process described in Section 3.1, annotators test several samples for all sub-tasks using ChatGPT-4o with a chain-of-thought (CoT) prompt to generate answers. They then determine whether each sub-task is reasoning-specific and thus should be retained. In addition, since math-specific visual reasoning benchmarks are well-established and relatively large in scale, we remove all hard-math questions to enable a more targeted evaluation of general multi-image reasoning scenarios. As illustrated in the left block of Figure \ref{fig:data-pipeline}, all remaining tasks form a reasoning-specific task pool containing 101 potential task candidates that are ready for the further annotation process.

\subsubsection{\textbf{Reasoning Steps Annotation}}

To construct annotated reasoning trajectories for a given task $T$, we begin with a set of input images $I$, a corresponding question $Q$, and the ground-truth final answer $G$. These elements $(I, Q, G)$ are provided as input to GPT-4o, which is prompted to generate three distinct solution paths $R = \{R_1, R_2, R_3\}$, each representing a different reasoning trajectory for arriving at the correct answer. Each path $R_i$ is composed of a sequence of reasoning steps $S_i$, where every step is designed to reflect a specific cognitive operation. Inspired by the classification scheme proposed in \textit{LLaVA-o1} \cite{xu2024llava-o1} and \textit{LLamaV-o1} \cite{thawakar2025llamav-o1} for intermediate reasoning annotation, we refine the categorization of these steps into six granular types: \textbf{Task Understanding}, \textbf{Information Grounding}, \textbf{Commonsense Seeking}, \textbf{Logical Reasoning}, \textbf{Arithmetic Calculating}, and \textbf{Draw Conclusion}. Each step produced by GPT-4o is required to conform to one of these predefined reasoning types. Through this process, the resulting annotated task can be formally represented as $T = (I, Q, G, R)$, where $R$ encapsulates multi-path, type-constrained reasoning trajectories aligned with the final answer.

\subsubsection{\textbf{Manual Inspection and Correction}}
\input{latex/tables/overall-results}

Although GPT-4o is one of the most powerful and balanced models available, it is not infallible. Therefore, a process of double-checking and correction by human annotators is applied to ensure the high quality of our benchmark. We recruit annotation volunteers to review the questions and ground-truth answers of the tasks, and to verify whether the reasoning steps generated by GPT-4o are consistent with the image content and the provided ground truth. Annotators are instructed to manually correct any inaccuracies found in the reasoning steps or the ground-truth answers from the data sources.

This annotation process involved approximately 200 working hours by 17 annotators, all of whom were at least undergraduate students majoring in computer science or artificial intelligence. In total, 1,198 out of 4,750 samples (about \textbf{25\%}) are corrected by at least one reasoning step. In addition, 355 final answers (about \textbf{7.5\%} of the total) that were initially incorrect in the data sources are identified and corrected.

\subsection{Subset for Multi-image Reward Model}

Taking advantage of the data annotation process of MMRB, we construct a subset of 2,313 samples to evaluate the capability of the multimodal reward model in multi-image scenarios. Based on the reasoning steps annotated by GPT-4o and the corrected reasoning steps provided by human experts, we label the incorrect annotations as rejected samples and the corrected ones as accepted samples. Each reward evaluation sample thus consists of a \textbf{Question}, \textbf{Answer}, \textbf{Multiple-images}, and both \textbf{Accepted and Rejected annotations}. To further evaluate the stability of the reward models, we construct two test sets with randomly ordered accepted and rejected samples, forming \textbf{Test Set 1} and \textbf{Test Set 2}.

%% file: latex/tables/dataset-statistics.tex
\begin{table}[h]
\begin{center}
\small
\begin{tabular}{llll}
\hline
Statistics        & Number & Statistics                & Number \\ \hline
Sub-Tasks         & 92     & Total Reasoning Steps     & 68,882 \\
---Multiple-choice & 73     & Avg Reasoning Steps       & 4.83   \\
---Free-form       & 19     & Avg Solutions per         & 1.93   \\
Total Samples     & 4,750  & Avg Question Length       & 492.67 \\
Total Images      & 29,293 & Corrected Ground Truth    & 355    \\
Avg Images        & 6.17   & Corrected Reasoning Paths & 1,198  \\ \hline
\end{tabular}
\end{center}
\caption{Statistics of MMRB benchmark.}
\label{dataset-statistics}
\end{table}



%% file: latex/tables/overall-results.tex
\begin{table*}[t]
\begin{center}
\small
\begin{tabular}{cccccccccc}
\hline
\textbf{\begin{tabular}[c]{@{}c@{}}Baseline\\ Model\end{tabular}} & \begin{tabular}[c]{@{}c@{}}Outcome\\ Score\\ (\%)\end{tabular} & \begin{tabular}[c]{@{}c@{}}Outcome\\ Score\\ w/ CoT (\%)\end{tabular} & \begin{tabular}[c]{@{}c@{}}Process\\ Score\\ (\%)\end{tabular} & \begin{tabular}[c]{@{}c@{}}Efficacy\\ Score\\ (\%)\end{tabular} & \textbf{\begin{tabular}[c]{@{}c@{}}Baseline\\ Model\end{tabular}} & \begin{tabular}[c]{@{}c@{}}Outcome\\ Score\\ (\%)\end{tabular} & \begin{tabular}[c]{@{}c@{}}Outcome\\ Score\\ w/ CoT (\%)\end{tabular} & \begin{tabular}[c]{@{}c@{}}Process\\ Score\\ (\%)\end{tabular} & \begin{tabular}[c]{@{}c@{}}Efficacy\\ Score\\ (\%)\end{tabular} \\ \hline\hline
\multicolumn{10}{c}{Non-reasoning-specialized   API Model}                                                                                                                                                                                                                                                                                                                                                                                                                                                                                                                                                                                                                                    \\ \hline
GPT-4o-20241120                                                   & 57.88                                                          & 64.04                                                                 & 83.70                                                          & 6.16                                                            & GPT-4o-mini-20240718                                              & 47.71                                                          & 57.92                                                                 & 79.90                                                          & 10.21                                                           \\
Claude-3.7-sonnet-0219                                            & 62.56                                                          & 64.33                                                                 & 81.88                                                          & 1.77                                                            & Gemini-2.5-flash-0417                                             & 62.61                                                          & 69.95                                                                 & 82.24                                                          & 7.34                                                            \\ \hline\hline
\multicolumn{10}{c}{Reasoning-specialized   API Model}                                                                                                                                                                                                                                                                                                                                                                                                                                                                                                                                                                                                                                        \\ \hline
GPT-o1-20241217                                                   & 72.04                                                          & 73.36                                                                 & 79.31                                                          & 1.32                                                            & GPT-o4-mini-20250416                                              & 69.29                                                          & 71.23                                                                 & 80.01                                                          & 1.94                                                            \\
Claude-3.7-sonnet-think                                           & 62.07                                                          & 65.98                                                                 & 85.76                                                          & 3.91                                                            & Gemini-2.5-flash-think                                            & 71.53                                                          & 71.43                                                                 & 82.13                                                          & -0.10                                                           \\
grok-3-think                                                      & 44.68                                                          & 47.16                                                                 & 84.41                                                          & 2.48                                                            & Gemini-2.5-pro-exp-0325                                           & 73.86                                                          & 71.38                                                                 & 89.78                                                          & -2.48                                                           \\ \hline\hline
\multicolumn{10}{c}{Non-reasoning-specialized   Open-source Model}                                                                                                                                                                                                                                                                                                                                                                                                                                                                                                                                                                                                                            \\ \hline
Qwen2VL-2B                                                        & 39.51                                                          & 36.30                                                                 & 3.32                                                           & -3.21                                                           & Qwen2VL-7B                                                        & 56.57                                                          & 54.50                                                                 & 5.77                                                           & -2.07                                                           \\
Qwen2.5VL-3B                                                      & 46.99                                                          & 45.15                                                                 & 46.12                                                          & -1.84                                                           & Qwen2.5VL-7B                                                      & 56.03                                                          & 56.66                                                                 & 65.87                                                          & 0.63                                                            \\
Qwen2.5VL-32B                                                     & 64.45                                                          & 62.49                                                                 & 79.92                                                          & -1.96                                                           & Qwen2.5-Omni-3B                                                   & 50.01                                                          & 50.44                                                                 & 9.78                                                           & 0.43                                                            \\
Qwen2.5-Omni-7B                                                   & 50.11                                                          & 46.51                                                                 & 17.74                                                          & -3.60                                                           & InternVL2.5-1B                                                    & 31.97                                                          & 33.00                                                                 & 12.65                                                          & 1.03                                                            \\
InternVL2.5-2B                                                    & 37.12                                                          & 37.60                                                                 & 12.20                                                          & 0.48                                                            & InternVL2.5-8B                                                    & 44.17                                                          & 42.78                                                                 & 49.84                                                          & -1.39                                                           \\
InternVL2.5-26B                                                   & 47.87                                                          & 47.44                                                                 & 22.60                                                          & -0.43                                                           & InternVL2.5-38B                                                   & 52.67                                                          & 55.35                                                                 & 59.19                                                          & 2.68                                                            \\
InternVL3-1B                                                      & 30.33                                                          & 26.25                                                                 & 37.16                                                          & -4.08                                                           & InternVL3-2B                                                      & 38.61                                                          & 38.58                                                                 & 39.65                                                          & -0.03                                                           \\
InternVL3-8B                                                      & 46.66                                                          & 48.76                                                                 & 59.74                                                          & 2.10                                                            & InternVL3-9B                                                      & 47.82                                                          & 48.81                                                                 & 35.27                                                          & 0.99                                                            \\
InternVL3-14B                                                     & 56.44                                                          & 58.50                                                                 & 65.48                                                          & 2.06                                                            & InternVL3-38B                                                     & 54.27                                                          & 58.72                                                                 & 70.93                                                          & 4.45                                                            \\
LLaVA-OneVision-0.5B                                              & 35.12                                                          & 33.01                                                                 & 3.81                                                           & -2.11                                                           & LLaVA-OneVision-7B                                                & 47.33                                                          & 43.99                                                                 & 7.86                                                           & -3.34                                                           \\
MiniCPM-V-2.6-8B                                                  & 50.19                                                          & 46.48                                                                 & 60.42                                                          & -3.71                                                           &                                                                   &                                                                &                                                                       &                                                                &                                                                 \\ \hline\hline
\multicolumn{10}{c}{Reasoning-specialized   Open-source Model}                                                                                                                                                                                                                                                                                                                                                                                                                                                                                                                                                                                                                                \\ \hline
R1-OneVision-7B-RL                                                & 52.93                                                          & 52.29                                                                 & 69.90                                                          & {\color[HTML]{333333} -0.64}                                    & Skywork-R1V2-38B                                                  & 45.21                                                          & 45.59                                                                 & 77.50                                                          & 0.38                                                            \\
QVQ-72B-Preview                                                   & 52.36                                                          & 51.47                                                                 & 82.38                                                          & {\color[HTML]{333333} -0.89}                                    &                                                                   &                                                                &                                                                       &                                                                &                                                                 \\ \hline
\end{tabular}
\end{center}
\caption{Overall results of 34 baseline models on MMRB benchmark. * indicates 1/5 data is applied for saving computing cost.}
\label{overall-results}
\end{table*}

%% file: latex/05-Evaluation-Metrics.tex
\section{Evaluation Metrics}
\label{evaluation-metrics}

\subsection{Rule-based Evaluation Metrics}

Accuracy are used as metrics for deterministic evaluation, taking advantage of the fact that the samples in MMRB are either multiple-choice or free-form questions. The evaluation protocol requires the model to generate a final answer in an extractable format, either as a direct answer or through step-by-step reasoning prompts. A rule-based answer extractor has been developed, with which an average of 96\% of the answers can be correctly identified, enabling direct calculation of the \textbf{outcome score}. For the evaluation of reward models, we calculate the ranking accuracy in a single pass (\textbf{Acc@1}) on the two test subsets.

\subsection{LLM-based Evaluation Metrics}

To automatically and precisely evaluate the intermediate reasoning steps of MLLMs, powerful LLMs appear to be a promising alternative to hiring human annotators. Following the initial attempt of previous works such as MME-CoT \cite{jiang2025mme-cot}, which evaluate the quality and efficiency of CoT answers using GPT-4o, we further develop an evaluation protocol with improved cost efficiency and interpretability, as the right block of Figure \ref{fig:data-pipeline} shows. Specifically, we applied Qwen3-32B \cite{qwen3}, which surpasses GPT-4o on OpenCompass \cite{2023opencompass} in the majority of subjects and is much cheaper and faster. Similar but not identical to \citet{jiang2025mme-cot}'s method, we determine the metrics based on direct sentence-level matching rather than first dividing them into steps. Finally, we report the \textbf{process score} and \textbf{efficacy score} of the CoT outputs for all involved MLLMs. See Appendix \ref{appx:llm-evaluation-metrics} for the detailed formula.

%% file: latex/06-Experimental-Setup.tex
\section{Experimental Setup}
\label{experimental-setup}

\subsection{Baseline Models}

We select 40 MLLMs for our baseline experiment, including 10 commercial models (e.g. GPT-o1\cite{jaech2024openai-o1}, Gemini-2.5-pro\cite{anil2023gemini}) and 24 open-source models ranging from 0.5B to 38B parameters, among which 9 are specifically designed for reasoning. 8 multimodal reward models are also involved. A detailed model categorization is provided in Appendix \ref{appx:baseline-models}. Models that do not support multi-image input (e.g., LLaVA-o1) or exhibited usability issues are excluded. Please see Appendix \ref{appx:implementation-details} for implementation details.


\subsection{Computational Budget}

We perform all inference on 8 NVIDIA A800-80GB GPUs using the NVIDIA CUDA platform. According to the API platforms, evaluating the commercial models for the baseline costs approximately \$1,344, while generating data annotations with GPT-4o costs approximately 25.2 million tokens.


%% file: latex/tables/reward-model-results.tex
\begin{table}[t]
\begin{center}
\small
\begin{tabular}{ccc}
\hline
\textbf{Baseline Model} & \begin{tabular}[c]{@{}c@{}}Test Set 1\\ Acc@1 (\%)\end{tabular} & \begin{tabular}[c]{@{}c@{}}Test Set 2 \\ Acc@1 (\%)\end{tabular} \\ \hline
GPT-4o                  & 45.51                                                            & 19.34                                                             \\
GPT-4o-mini             & 85.69                                                            & 10.20                                                             \\
Gemini-2.0              & 47.29                                                            & 46.56                                                             \\ \hline
LLaVA-Critic-7B         & 54.02                                                            & 30.01                                                             \\
VisualPRM-8B            & 32.43                                                            & 46.70                                                             \\
R1-Reward-8B            & 29.58                                                            & 80.20                                                             \\
Skywork-VL-Reward-7B\textsuperscript{*}    & 38.00                                                            & 38.00                                                             \\
IXC-2.5-Reward-7B\textsuperscript{*}       & 39.04                                                            & 39.04                                                             \\ \hline
\end{tabular}
\end{center}
\caption{Results of 8 multimodal reward models on MMRB reward subset. * indicates score output mode.}
\label{reward-model-results}
\end{table}

%% file: latex/07-Main-Results-and-Discussion.tex
\section{Results and Discussion}
\label{results-and-discussion}

\subsection{Overall Quantitative Results}
We present the overall performance of 34 baseline models in Table \ref{overall-results}, including outcome scores, process scores, and efficacy scores. 
The MMRB benchmark remains highly challenging for open-source MLLMs, with average outcome and process scores below 50\%. Commercial models perform better but are not flawless, averaging 65\% in outcome and 83\% in process accuracy. The key findings are summarized as follows:


\subsubsection{\textbf{Inference-time scaling is effective for multi-image reasoning tasks}}

When comparing reasoning-specific models with their non-reasoning-specific counterparts (e.g., GPT-4o vs. GPT-o1), reasoning-specific API models demonstrate a significant performance advantage, with an average 6\% improvement in outcome scores over their non-reasoning counterparts, while their reasoning process scores remain roughly the same on average.

\subsubsection{\textbf{Even for reasoning-specific models, explicit CoT thinking can lead to performance improvements, though the gains are more moderate compared with non-reasoning models}}

This tendency is particularly evident in API models, where CoT prompting yields an average 6.4\% performance improvement for non-reasoning models, compared to a more moderate 1.2\% for reasoning-specific models.

\subsubsection{\textbf{Model scaling is validated for multi-image reasoning tasks}}

When comparing different versions within the same model family (e.g., Qwen2VL vs. Qwen2.5VL), improvements in data scale and base model capability consistently lead to better performance—yielding an average 5\% increase in outcome scores for the Qwen series and 3\% for the InternVL series, along with a significant 20\% improvement in process scores for the InternVL series. On the other hand, within each open-source model family (e.g., InternVL3 series), performance also generally improves as model size increases. On the other hand, although published close in time, the performance gap across different model families still exists, potentially due to variations in training strategies and datasets.

\subsubsection{\textbf{For non-reasoning-specific models, step-by-step thinking reveals benefits for models larger than 8B}}

By analyzing the efficacy score—defined as the difference in accuracy between CoT-prompted and non-CoT responses—a positive correlation is observed between model size and CoT efficacy. On average, open-source models larger than 8B achieve an efficacy score of 1.3, and API models average a score of 3.3, whereas smaller models average -1.4. As shown in Figure \ref{fig:model-size-vs-efficacy}, for every 1B increase in model size, the efficacy score increases by approximately 0.095 points. This effect is statistically significant (p = 0.028), and explains about 23\% of the variance (R² = 0.231).

\subsubsection{\textbf{Instruction following has a significant impact on the reasoning process score}}

An inspection of CoT answers reveals that smaller open-source models are less likely to follow instructions to produce answers in an extractable format or to explicitly generate step-by-step reasoning. As a result, models like Qwen2VL-2B and LLaVA-OneVision-0.5B tend to receive low reasoning process scores. In addition, omni-models like Qwen2.5-Omni appear to overfit on voice-chatting data, and therefore often fail to generate reasoning steps in most cases.


\subsection{Multimodal Reward Model Evaluation}

Table \ref{reward-model-results} shows the ranking accuracy of three API models and five open-source multimodal reward models. These reward models achieve an average ranking accuracy of 46.5\% on Test Set 1 with a single pass and 38.8\% on Test Set 2.

Interestingly, simply reversing the order of accepted and rejected samples can significantly impact performance. For example, GPT-4o-mini scores 85.69\% on Test Set 1 but drops to 10.20\% on Test Set 2. Other models show similar trends, except Skywork-VL-Reward and IXC-2.5-Reward, which only output reward scores rather than comparisons. Only Gemini-2.0 seems stable in this task.

This suggests that multi-image scenarios present a significant challenge for multimodal reward models, especially since most of them lack training data specific to multi-image inputs. This observation warrants further investigation in future research.


%% file: latex/figures/model-size-vs-efficacy.tex
\begin{figure}[t]
  \centering
  \includegraphics[width=\linewidth]{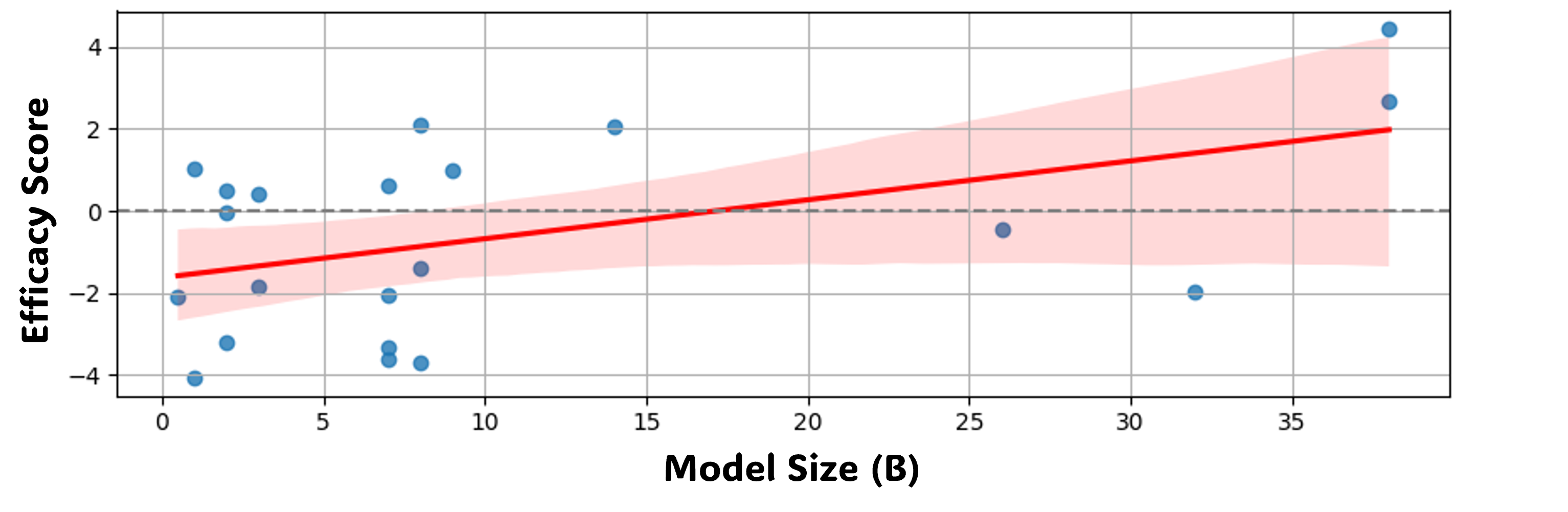}
  \caption{Linear regression on non-thinking models' efficacy.}
  \label{fig:model-size-vs-efficacy}
\end{figure}

%% file: latex/08-Conclusion.tex
\section{Conclusion}
\label{conclusion}

This paper contributes to the field of multimodal multi-image reasoning by introducing a challenging new benchmark, MMRB. Through extensive data collection and expert annotation, it presents 92 multi-image-only sub-tasks comprising over 68,000 reasoning steps. MMRB is the first benchmark of its kind designed for general-purpose multi-image reasoning scenarios. Comprehensive baseline evaluations are conducted using 40 API and open-source MLLMs, leading to several important findings. The results suggest that current open-source models exhibit limited multi-image reasoning capabilities, and multimodal reward models hold significant potential for future research.

%% file: latex/Appendix/_Appendix-Overview.tex
\appendix
\section*{Appendix Overview}

\noindent \textbf{Appendix \ref{appx:minimum-sample-capacity}}: Minimum Sample Capacity Estimation

\noindent \textbf{Appendix \ref{appx:llm-evaluation-metrics}}: LLM-based Evaluation Metrics

\noindent \textbf{Appendix \ref{appx:implementation-details}}: Implementation Details

\noindent \textbf{Appendix \ref{appx:error-analysis}}: Error Analysis

\noindent \textbf{Appendix \ref{appx:data-samples}}: Data Samples

\noindent \textbf{Appendix \ref{appx:data-sources}}: Data Sources

\noindent \textbf{Appendix \ref{appx:baseline-models}}: Baseline Models

\noindent \textbf{Appendix \ref{appx:prompt-templates}}: Prompt Templates

%% file: latex/Appendix/A-Minimum-Sample-Capacity-Estimation.tex
\section{Minimum Sample Capacity Estimation}
\label{appx:minimum-sample-capacity}

To optimize the cost efficiency of our benchmark, we estimate the minimum required sample size for each sub-task through a pilot experiment. This experiment evaluates how the stability and statistical significance of test results vary with different sample sizes.

Three distinct sub-tasks are selected, each involving multi-image contexts and focusing on a specific skill: reading comprehension (Slide QA), temporal understanding (Action Recognition), or abstract reasoning (IQ test). Using the data construction pipeline described in Section 4, 200 samples are annotated with detailed reasoning steps. Six baseline models ranging from 1B to 32B parameters are evaluated, and their performance is analyzed in terms of both outcomes and reasoning processes, as discussed in Section 5.

We use the Standard Error of Mean (SEM) of accuracy as an indicator of testing stability. Let \( X_1, X_2, \ldots, X_n \) be \( n \) independent samples drawn from the distribution of a certain performance metric, with the sample mean defined as:
\[
\bar{X} = \frac{1}{n} \sum_{i=1}^{n} X_i
\]
The standard error of the mean (SEM) is given by:
\[
\text{SEM} = \frac{\sigma}{\sqrt{n}}
\]
where \( \sigma \) is the population standard deviation. Since \( \sigma \) is typically unknown in practice, we use the sample standard deviation \( s \) to estimate SEM:
\[
\widehat{\text{SEM}} = \frac{s}{\sqrt{n}}, \quad \text{where} \quad s = \sqrt{ \frac{1}{n-1} \sum_{i=1}^{n} (X_i - \bar{X})^2 }
\]
In our experimental setup, we aim for a 95\% confidence level and require the width of the confidence interval for the estimated mean to be within 4\%. Based on the confidence interval formula under the normality assumption:
\[
\bar{X} \pm z_{0.025} \cdot \widehat{\text{SEM}}, \quad \text{with} \quad z_{0.025} = 1.96
\]
To satisfy this constraint, the upper bound of the SEM must meet:
\[
\widehat{\text{SEM}} < \frac{0.04}{2 \cdot 1.96} \approx 0.0102
\]
Empirically, we perform the following procedure: 

For a given sample size n, randomly drawn from the full set of 200 samples, the SEM is estimated by performing 100 repeated samplings. This experiment is conducted for sample sizes ranging from 1 to 200, in increments of one. A 95\% confidence level is used to estimate the minimum sample size required to achieve sufficiently stable results for each sub-task.

As shown in the SEM-versus-sample-size curve in Figure \ref{fig:sem-vs-n-outcome} and Figure \ref{fig:sem-vs-n-process}, the minimum sample size satisfying our stability condition is 18 for the outcome score and 10 for the process score. To ensure robust evaluation, we ultimately set a uniform sample size of 50, which lies well within the safe zone.

\input{latex/figures/sem-vs-n-outcome}
\input{latex/figures/sem-vs-n-process}


%% file: latex/figures/sem-vs-n-outcome.tex
\begin{figure}[h]
  \centering
  \includegraphics[width=\linewidth]{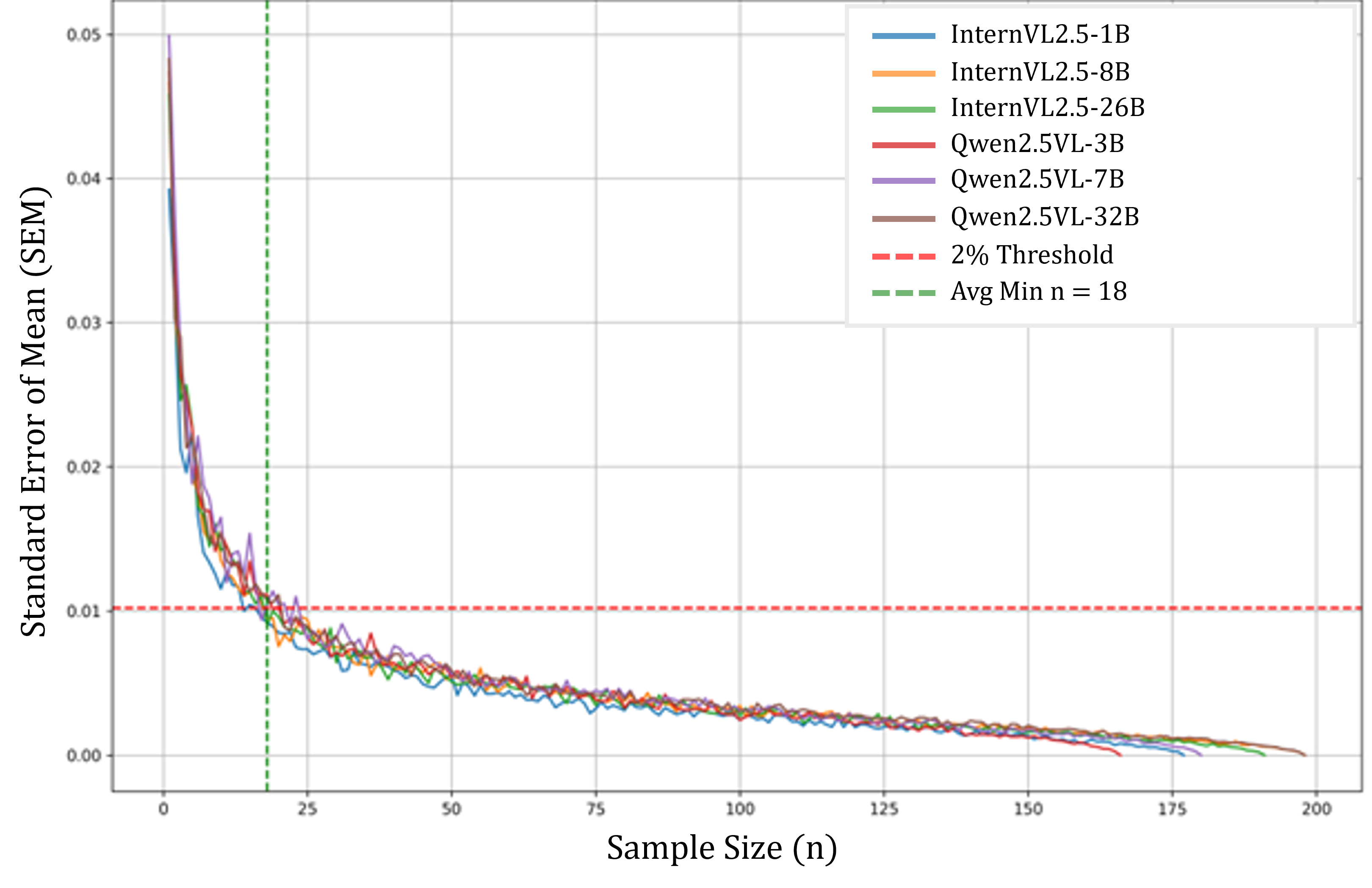}
  \caption{Standard Error of Mean (SEM) VS Sample Size (n) for outcome score.}
  \label{fig:sem-vs-n-outcome}
\end{figure}

%% file: latex/figures/sem-vs-n-process.tex
\begin{figure}[h]
  \centering
  \includegraphics[width=\linewidth]{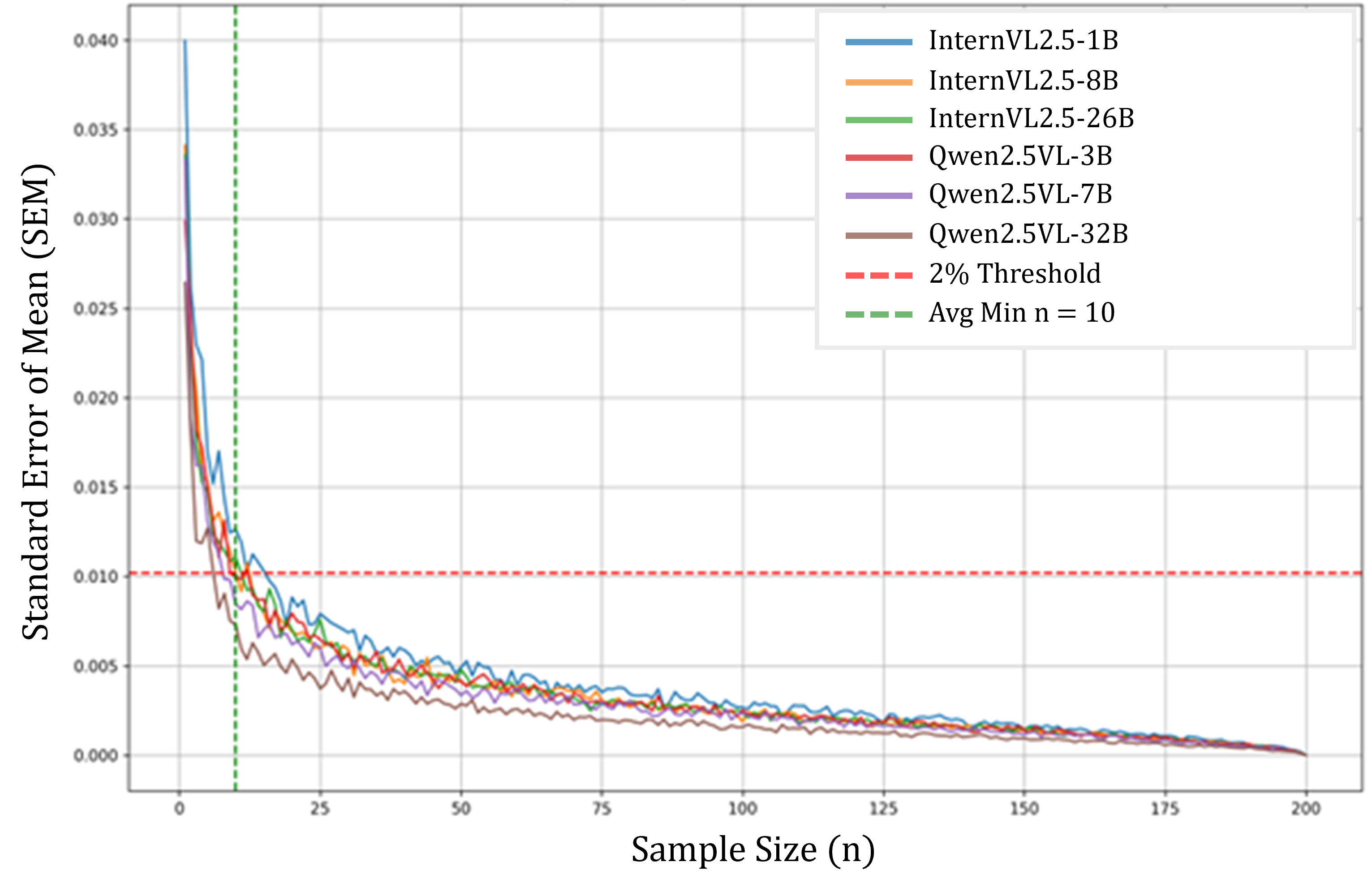}
  \caption{Standard Error of Mean (SEM) VS Sample Size (n) for process score..}
  \label{fig:sem-vs-n-process}
\end{figure}

%% file: latex/Appendix/B-Evaluation-Metrics.tex
\section{LLM-based Evaluation Metrics}
\label{appx:llm-evaluation-metrics}

\subsection{\textbf{Precision Score}}

Given a CoT answer \( C \), which consists of a set of phrases \( P = \{p_1, p_2, \dots, p_n\} \), we evaluate its quality by comparing each annotated reasoning step \( s_i \in R_i \) from the ground-truth trajectory with the phrases in \( P \).

A reasoning step \( s_i \) is considered \textbf{correct} if there exists a phrase \( p_j \in P \) such that \( p_j \) matches \( s_i \) at the sentence level---demonstrating either lexical or paraphrastic similarity---and simultaneously conveys the same cognitive operation as defined by the reasoning type of \( s_i \). Otherwise, \( s_i \) is marked as incorrect.

The \textbf{Precision} of a reasoning trajectory \( R_i = \{s_1, s_2, \dots, s_k\} \) is defined as the proportion of correctly recovered steps:
\[
\text{Precision}(R_i) = \frac{1}{k} \sum_{i=1}^{k} \mathbb{1}\left[\exists\, p_j \in P \text{ such that } \text{Match}(p_j, s_i)\right],
\]
where \( \mathbb{1}[\cdot] \) is the indicator function and \( \text{Match}(p_j, s_i) \) denotes both sentence-level and semantic (reasoning-type) alignment between \( p_j \) and \( s_i \).

Given a task \( T \), where \( R = \{R_1, R_2, R_3\} \) denotes three candidate reasoning trajectories, the \textbf{process score} for the task is defined as:
\[
\text{Process Score}(T) = \max_{R_i \in R} \text{Precision}(R_i).
\]

\subsection{\textbf{Efficacy Score}}

We follow the exact efficacy calculation used in MME-CoT \cite{jiang2025mme-cot}. This allows us to evaluate how incorporating step-by-step thinking affects reasoning accuracy, where:
\[
\text{Efficacy} = \text{AccR}_{\text{COT}} - \text{AccR}_{\text{DIR}}
\]





%% file: latex/Appendix/C-Implementation-Details.tex
\section{Implementation Details}
\label{appx:implementation-details}

We implement the open-source model inference process using the Transformers library, following the officially recommended environment setup for each baseline model. For multi-image inputs, we adopt the recommended scaling and patching parameters from the models’ demo code. If the GPU runs out of memory for certain data entries, we reduce the image input to the minimum quality and rerun the inference. Commercial models are accessed through their official API platforms, and one-fifth of the data is used for expensive reasoning models to reduce costs. All data and the open-source models used are released under open-source licenses. 

The prompt for directly generating an answer is: “Do not include any explanation. Only output your final answer in the exact format: Answer[<letter> or <your\_answer\_here>]", and the prompt for generating a CoT answer is: “Please think step by step. Then write your final answer in the format: Answer[<letter> or <your\_answer\_here>].”  

%% file: latex/Appendix/D-Error-Analysis.tex
\section{Error Analysis}
\label{appx:error-analysis}

We analyze the error patterns in the MLLM's reasoning and answer. A total of 282 model responses from Open AI o1 and identified 6 distinct types of errors that harm the performance. As illustrates in Figure \ref{fig:error-analysis}, about a third of the errors result by incorrect reasoning, which is because of the model is un-optimiazed capability; another main error source comes from failing to grounding the information from the image or understanding wrong image messages; ans about 15\% of the errors come from failure to follow problem instructions. It is also noticable that about 18\% of the orrors are accetable ones where the answers can be considered correct by part of annotators but not identical with the ground truth. There also exists a small number of errors that is caused by stubbore adherence to the model it self's common sense, and failed to recognize the image order.

\input{latex/figures/error-analysis}

%% file: latex/figures/error-analysis.tex
\begin{figure}[h]
  \centering
  \includegraphics[width=\linewidth]{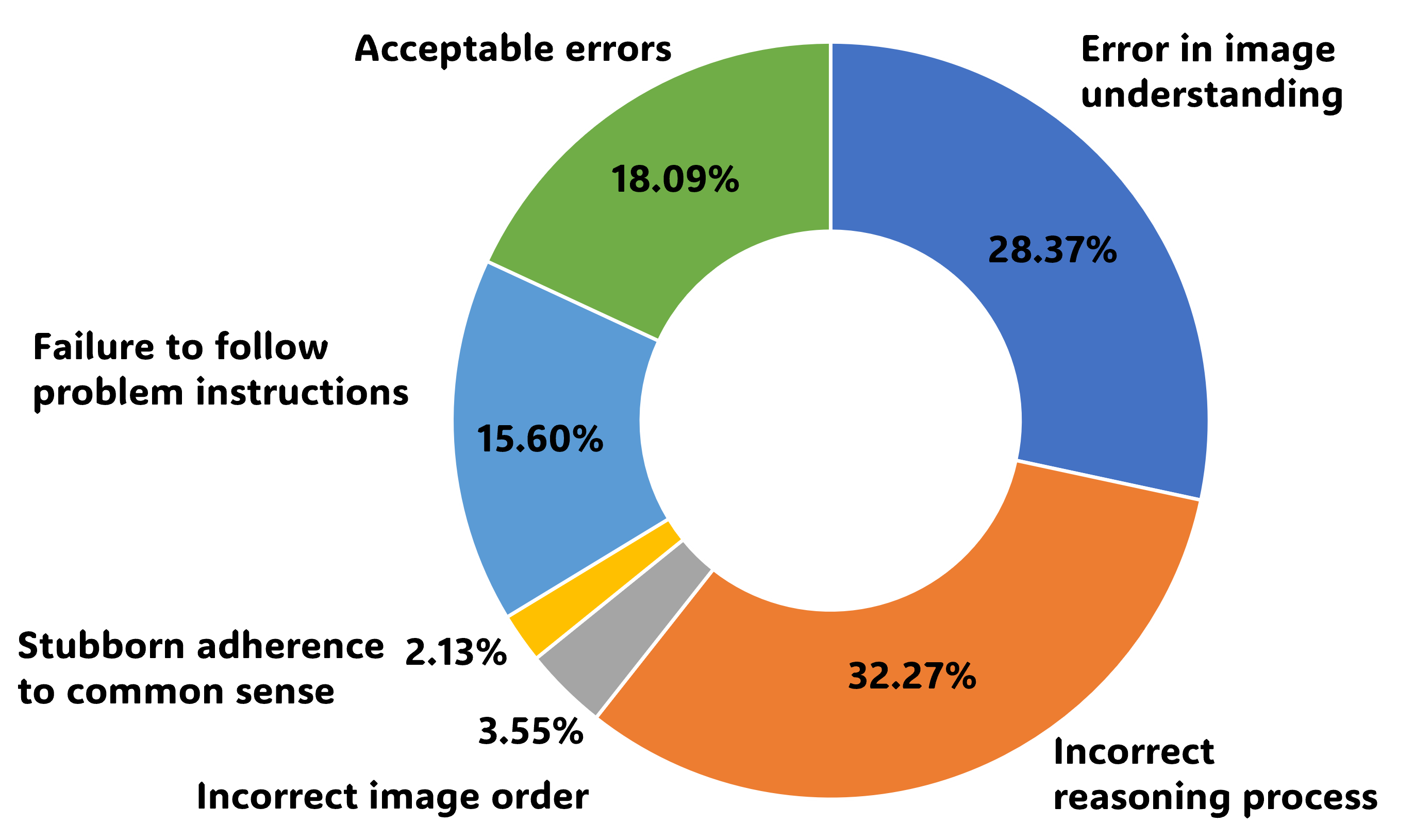}
  \caption{Error analysis of Open AI o1's model response.}
  \label{fig:error-analysis}
\end{figure}

%% file: latex/Appendix/E-Data-Samples.tex
\section{Data Samples}
\label{appx:data-samples}

Please see Figure \ref{fig:dataset-example-1} and Figure \ref{fig:dataset-example-2}

\input{latex/figures/dataset-example-1}
\input{latex/figures/dataset-example-2}

%% file: latex/figures/dataset-example-1.tex
\begin{figure*}[t]
  \centering
  \includegraphics[width=0.8\linewidth]{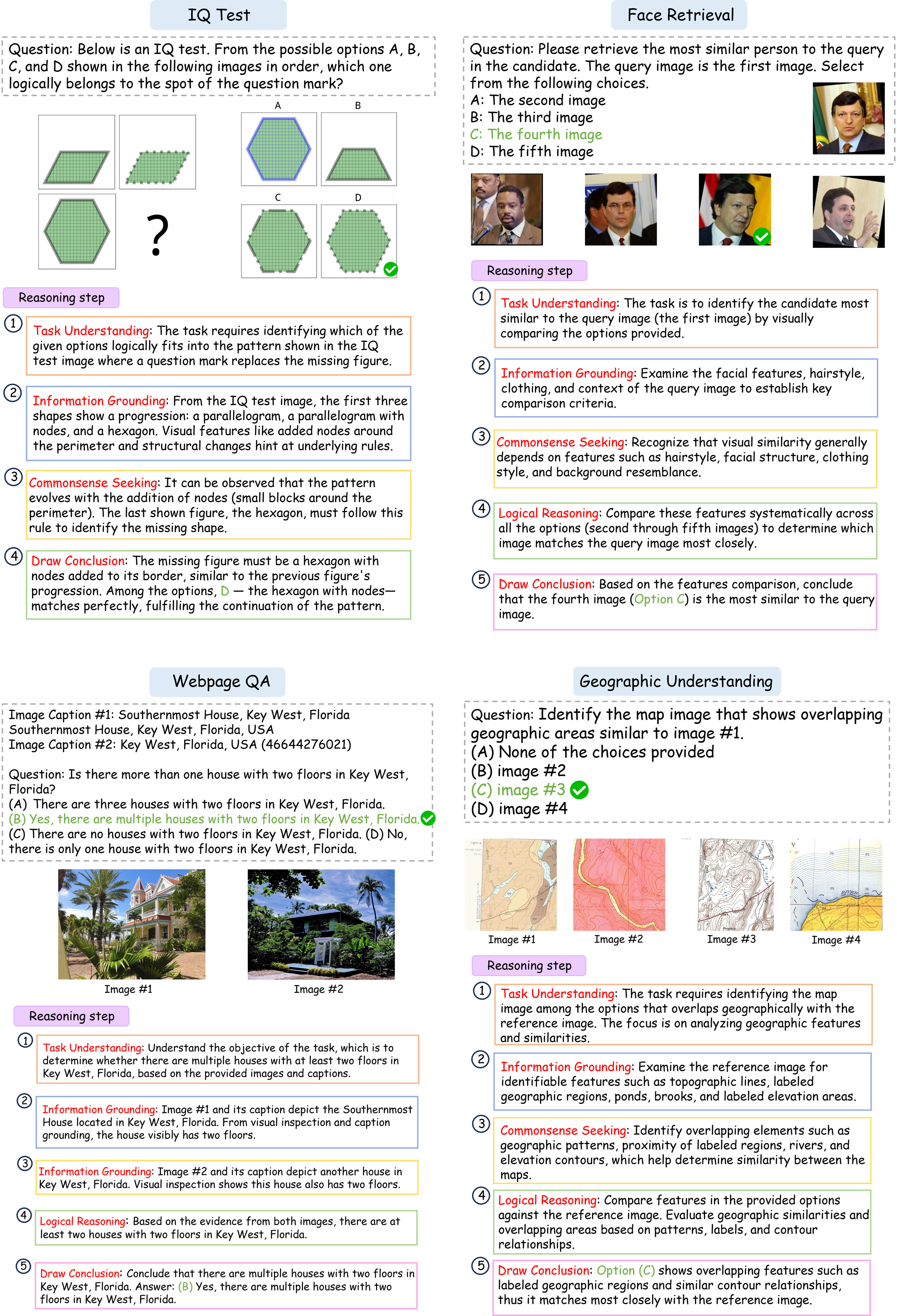}
  \caption{Examples of MMRB benchmark.}
  \label{fig:dataset-example-1}
\end{figure*}

%% file: latex/figures/dataset-example-2.tex
\begin{figure*}[t]
  \centering
  \includegraphics[width=0.8\linewidth]{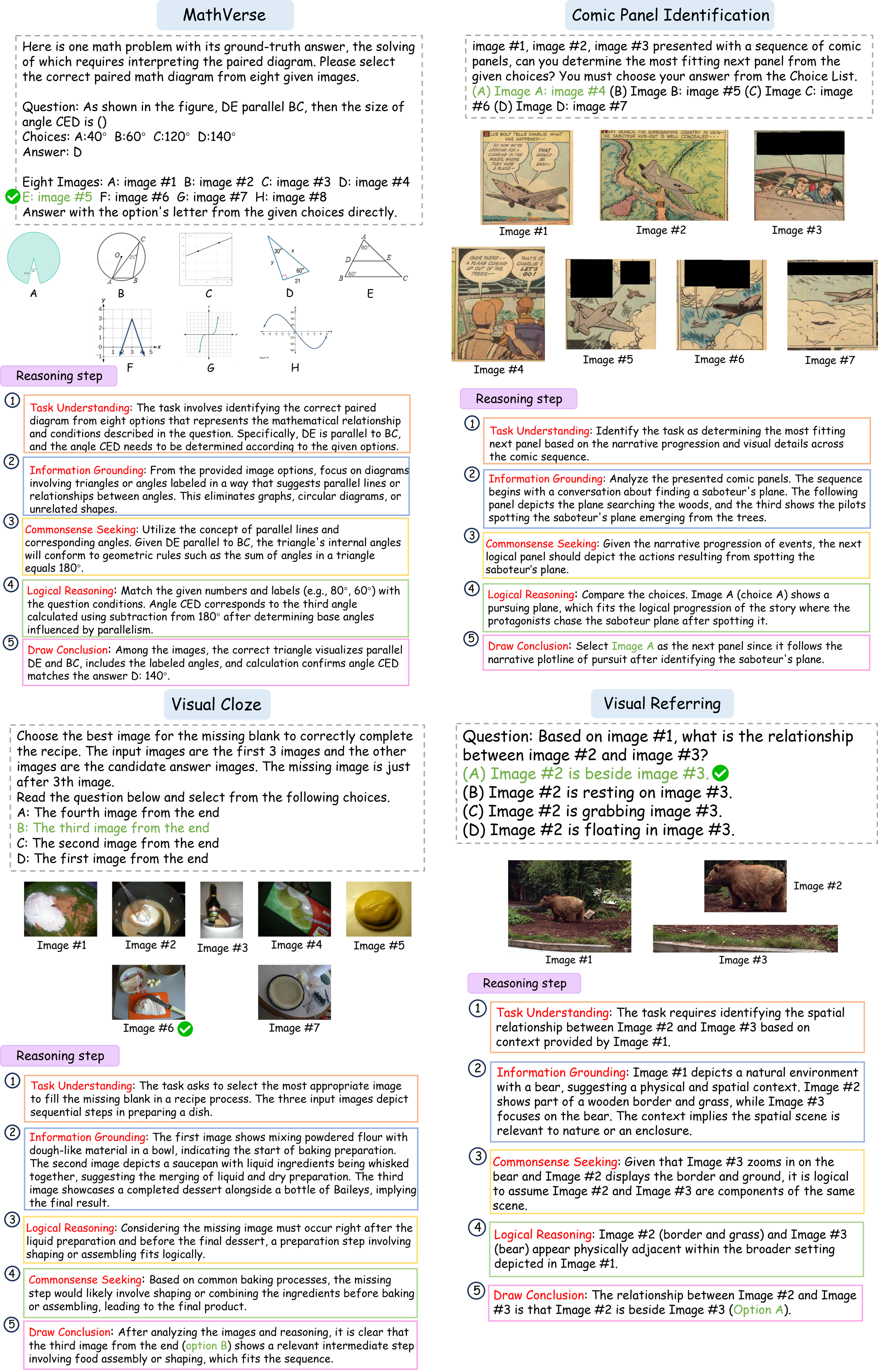}
  \caption{Examples of MMRB benchmark.}
  \label{fig:dataset-example-2}
\end{figure*}

%% file: latex/Appendix/F-Data-Sources.tex
\section{Data Sources}
\label{appx:data-sources}

Please see Table \ref{tab:dataset-sources}.
\input{latex/tables/dataset-sources}

%% file: latex/tables/dataset-sources.tex
\begin{table*}[t]
\begin{center}
\small
\begin{tabular}{ccccc}
\hline
Dataset                     & Publish Date       & Task                                                                                                                                                                      & Total Samples & Modality               \\ \hline
DEMON                       & May  25,2024       & Multimodal Dialogue, Multi-image understanding                                                                                                                            & 18.18K        & Text \& Image          \\
LLaVA-NeXT-Interleave-Bench & June  16,2024      & \begin{tabular}[c]{@{}c@{}}Multi-image, Multi-frame, \\ Multi-view, Multi-patch\end{tabular}                                                                              & 1,177.6K      & Text \& Image \& Video \\
Mantis-Eval                 & June  18,2024      & \begin{tabular}[c]{@{}c@{}}Co-reference, Comparison, Reasoning, \\ Temporal understanding\end{tabular}                                                                    & 721K          & Text \& Image \& Video \\
MIBench                     & August  21,2024    & Multi-image understanding, ICL                                                                                                                                            & 9.6K          & Text \& Image          \\
MileBench                   & April  29,2024     & Temporal Multi-Image, Semantic Multi-Image                                                                                                                                & 6.4K          & Text \& Image \& Video \\
MIRB                        & June  19,2024      & \begin{tabular}[c]{@{}c@{}}Multi-Image  Reasoning, Visual World Knowledge, \\ Perception, Multi-Hop Reasoning\end{tabular}                                                & 0.9K          & Text \& Image          \\
MMlU-Benchmark              & September  23,2024 & Multi-image understanding                                                                                                                                                 & 11.6K         & Text \& Image          \\
MUIRBENCH                   & June  15,2024      & Multi-image understanding                                                                                                                                                 & 2.6K          & Text \& Image          \\
ReMI                        & June  13,2024      & \begin{tabular}[c]{@{}c@{}}Mathematical reasoning, Spatial reasoning, \\ Logical reasoning\end{tabular}                                                                   & 2.6K          & Text \& Image          \\
SEED-Bench-2                & November  28,2024  & \begin{tabular}[c]{@{}c@{}}Multi-image and text understanding, \\ Interlaced graphic and text understanding, \\ Image generation and graphic-text generation\end{tabular} & 24.3K         & Text \& Image          \\ \hline
\end{tabular}
\end{center}
\caption{The data sources of MMRB benchmark.}
\label{tab:dataset-sources}
\end{table*}

%% file: latex/Appendix/G-Baseline-Models.tex
\section{Baseline Models}
\label{appx:baseline-models}

Please see Table \ref{appx:baseline-models}
\input{latex/tables/baseline-models}

%% file: latex/tables/baseline-models.tex
\begin{table*}[t]
\begin{center}
\small
\begin{tabular}{cccccc}
\hline
Model                                    & Size                       & Context Length & thinking & Tokens per Image & Multi-Image Training \\ \hline
GPT-4o-20241120                          & $\approx$200B                      & 128K           & no       & $\approx$170–340         & yes                  \\
GPT-4o-mini-20240718                     & $\approx$8B                        & 128K           & no       & $\approx$170–340         & yes                  \\
Claude-3.7-sonnet-20250219               & $\approx$175B                      & 200K           & no       & 1600             & yes                  \\
Gemini-2.5-flash-preview-nothinking-0417 & $\approx$8–20B                     & 1M             & no       & $\approx$258–1,032       & yes                  \\
Gemini-2.0-flash                         & $\approx$100B                      & 1M             & no       & $\approx$258–1,032       & yes                  \\ \hline
GPT-o1-20241217                          & $\approx$300B                      & 128K           & yes      & $\approx$170–340         & yes                  \\
GPT-o4-mini-20250416                     & $\approx$8B                        & 128K           & yes      & $\approx$170–340         & yes                  \\
Claude-3.7-sonnet-thinking-20250219      & $\approx$175B                      & 200K           & yes      & 1600             & yes                  \\
Gemini-2.5-flash-preview-thinking-0417   & $\approx$8–20B                     & 1M             & yes      & $\approx$258–1,032       & yes                  \\
Gemini-2.5-pro-exp-0325                  & $\approx$100–200B                  & 1M             & yes      & $\approx$258–1,032       & yes                  \\
grok-3-think                             & $\approx$140B                      & 1M             & yes      & $\approx$1,600           & yes                  \\ \hline
Qwen2VL                                  & 2B, 7B                     & 32768          & no       & 4 - 16384        & yes                  \\
Qwen2.5VL                                & 3B, 7B, 32B                & 32768          & no       & 4 - 16384        & yes                  \\
Qwen2.5-Omni                             & 3B, 7B                     & 32768          & no       & 4 - 16384        & yes                  \\
InternVL2.5                              & 1B, 2B, 8B, 26B, 38B       & 16384          & no       & 256              & yes                  \\
InternVL3                                & 1B, 2B, 8B, 9B, 14B,   38B & 8192           & no       & 256              & yes                  \\
LLaVA-OneVision                          & 0.5B, 7B                   & 32768          & no       & 729              & yes                  \\
MiniCPM-V-2.6                            & 8B                         & 32768          & no       & 640              & yes                  \\ \hline
R1-OneVision-7B-RL                       & 7B                         & 32768          & yes      & 4 - 16384        & no                   \\
Skywork-R1V2                             & 38B                        & 131072         & yes      & 256              & no                   \\
QVQ-Preview                              & 72B                        & 32768          & yes      & 4 - 16384        & no                   \\ \hline
IXC-2.5-Reward                           & 7B                         & 24K            & no       & 256              & no                   \\
LLaVA-Critic                             & 7B                         & 32768          & no       & 729              & no                   \\
Skywork-VL-Reward                        & 7B                         & 32768          & no       & 256              & no                   \\
VisualPRM                                & 8B                         & 24K            & no       & 256              & no                   \\
R1-Reward                                & 8B                         & 32768          & no       & 4 - 16384        & no                   \\ \hline
\end{tabular}
\end{center}
\caption{Details of baseline models involved.}
\label{tab:baseline-models}
\end{table*}

%% file: latex/Appendix/H-Prompt-Templates.tex
\section{Prompt Templates}
\label{appx:prompt-templates}

\subsection{Reasoning Step Generation by GPT-4o}
Please see Figure~\ref{prompt-reasoning}.

\begin{figure*}[htbp]
    \centering
    \begin{tcolorbox}[colframe=red!8!white, colback=yellow!15!white, coltitle=black, title=Prompt Template for Reasoning Step Generation by GPT-4o]
    \textbf{Prompt:}\\

    \texttt{REASONING\_GENERATION\_PROMPT = """}\\
    You are an expert in multimodal reasoning. Given a multi-image reasoning task, generate three distinct reasoning paths in JSON format, where each path follows a step-by-step Chain of Thought (CoT).\\
    Each reasoning step must include:\\
    - reasoning step: The step index.\\
    - reasoning type: Categorized as one of the following:\\
    \quad Task Understanding / Information Grounding / Commonsense Seeking / Logical Reasoning / Arithmetic Calculating / Draw Conclusion\\
    - rationale: A detailed explanation of the reasoning process at each step.\\
    The output must be a JSON list of three reasoning paths, where each path is a list of step-by-step reasoning objects like this:\\
    \[\\
    \quad \{\\
    \quad\quad "reasoning step": int,\\
    \quad\quad "reasoning type": str,\\
    \quad\quad "rationale": str\\
    \quad \},\\
    \quad ...\\
    \]\\
    Question: \{question\}\\
    Options: [\{options\}]\\
    Answer: \{answer\}\\
    \texttt{"""}
    \end{tcolorbox}
    \caption{Prompt Template for Reasoning Step Generation by GPT-4o.}
    \label{prompt-reasoning}
\end{figure*}

\subsection{Reasoning Steps Evaluation by Qwen3-32B}
Please see Figure~\ref{prompt-evaluation}.

\begin{figure*}[htbp]
    \centering
    \begin{tcolorbox}[colframe=red!8!white, colback=yellow!15!white, coltitle=black, title=Prompt Template for Reasoning Steps Evaluation by Qwen3-32B]
    \textbf{Prompt:}\\

    \texttt{STEP\_CORRECTNESS\_PROMPT = """}\\
    You are an expert in evaluating the correctness of reasoning steps.\\
    I have a multi-image reasoning task. Below, I provide:\\
    1. A human annotated step-by-step solution to solving this task.\\
    2. A Chain-of-Thought (CoT) answer generated by an LLM to be evaluated.\\

    Your task:\\
    - Compare the CoT answer to the step-by-step solution.\\
    - Match the related sentences from the CoT answer to the corresponding steps in the solution.\\
    - For each reasoning step in the CoT answer, determine if it is correct or incorrect according to the matched sentences from CoT answer.\\
    - Return in JSON format:\\
    \quad\texttt{\{\\
    \quad\quad "reasoning step": int,\\
    \quad\quad "reasoning type": str,\\
    \quad\quad "matched sentences": <Corresponding sentence from CoT answer> / null,\\
    \quad\quad "correctness": true / false\\
    \quad\}}\\
    Input:\\
    \textbf{Step-by-step solution:}\\
    \{step\_solution\}\\
    \textbf{CoT Answer:}\\
    \{CoT\_a\}\\
    \texttt{"""}
    \end{tcolorbox}
    \caption{Prompt Template for Reasoning Steps Evaluation by Qwen3-32B.}
    \label{prompt-evaluation}
\end{figure*}

\subsection{Reward Model Evaluation for API Models}
Please see Figures~\ref{prompt-gemini}, \ref{prompt-gpt4o}, and \ref{prompt-gpt4omini}.

\begin{figure*}[htbp]
    \centering
    \begin{tcolorbox}[colframe=red!8!white, colback=yellow!15!white, coltitle=black, title=Prompt Template for Reward Model Evaluation – Gemini 2]
    \textbf{Prompt:}\\

    \texttt{"""\{prompt\_prefix\}}\\
    You are a visual language model evaluation assistant. Please compare the two responses to the same question based on the following four aspects, determine which one is better, and explain why:\\

    1. Accuracy of target description\\
    2. Accuracy of relationship description\\
    3. Accuracy of attribute description\\
    4. Usefulness (informativeness/helpfulness)\\

    Please strictly choose the better response.\\
    You must choose a better answer, you can't judge them as "equally good". Try your best to select a better one.\\
    Output in the following format:\\
    The better response: [1]. Because...\\
    or\\
    The better response: [2]. Because...\\

    \textbf{[Task Content]}\\
    Image and Question:\\
    \{sample.get('question', 'N/A')\}\\

    Answer:\\
    \{sample.get('answer', 'N/A')\}\\

    Response 1:\\
    \{modified\_steps\}\\

    Response 2:\\
    \{raw\_steps\}\\
    \texttt{"""}
    \end{tcolorbox}
    \caption{Prompt Template for Reward Model Evaluation – Gemini 2.}
    \label{prompt-gemini}
\end{figure*}

\begin{figure*}[htbp]
    \centering
    \begin{tcolorbox}[colframe=red!8!white, colback=yellow!15!white, coltitle=black, title=Prompt Template for Reward Model Evaluation – GPT-4o]
    \textbf{Prompt:}\\

    \texttt{"""}\\
    You are a visual language model evaluation assistant. Please compare the two responses to the same question based on the following four aspects, determine which one is better, and explain why:\\

    1. Accuracy of target description\\
    2. Accuracy of relationship description\\
    3. Accuracy of attribute description\\
    4. Usefulness (informativeness/helpfulness)\\

    Please strictly choose the better response. You must choose a better answer, you can't judge them as "equally good". Try your best to select a better one.\\
    Output in the following format:\\
    The better response: [1]. Because...\\
    or\\
    The better response: [2]. Because...\\

    \textbf{[Task Content]}\\
    Image and Question:\\
    \{sample.get('question', 'N/A')\}\\

    Answer:\\
    \{sample.get('answer', 'N/A')\}\\

    Response 1:\\
    \{modified\_steps\}\\

    Response 2:\\
    \{raw\_steps\}\\
    \texttt{"""}
    \end{tcolorbox}
    \caption{Prompt Template for Reward Model Evaluation – GPT-4o.}
    \label{prompt-gpt4o}
\end{figure*}

\begin{figure*}[htbp]
    \centering
    \begin{tcolorbox}[colframe=red!8!white, colback=yellow!15!white, coltitle=black, title=Prompt Template for Reward Model Evaluation – GPT-4o-mini]
    \textbf{Prompt:}\\

    \texttt{"""}\\
    You are a visual language model evaluation assistant. Please compare the two responses to the same question based on the following four aspects, determine which one is better, and explain why:\\

    1. Accuracy of target description\\
    2. Accuracy of relationship description\\
    3. Accuracy of attribute description\\
    4. Usefulness (informativeness/helpfulness)\\

    Please strictly choose the better response. You must choose a better answer, you can't judge them as "equally good". Try your best to select a better one.\\
    Output in the following format:\\
    The better response: [1]. Because...\\
    or\\
    The better response: [2]. Because...\\

    \textbf{[Task Content]}\\
    Image and Question:\\
    \{sample.get('question', 'N/A')\}\\

    Answer:\\
    \{sample.get('answer', 'N/A')\}\\

    Response 1:\\
    \{modified\_steps\}\\

    Response 2:\\
    \{raw\_steps\}\\
    \texttt{"""}
    \end{tcolorbox}
    \caption{Prompt Template for Reward Model Evaluation – GPT-4o-mini.}
    \label{prompt-gpt4omini}
\end{figure*}

\subsection{Reward Model Evaluation for Open-source Reward Models}
Please see Figures~\ref{prompt-llava} and \ref{prompt-r1}.

\begin{figure*}[htbp]
    \centering
    \begin{tcolorbox}[colframe=red!8!white, colback=yellow!15!white, coltitle=black, title=Prompt Template for Reward Model Evaluation – llava-critic]
    \textbf{Prompt:}\\

    """\{\{"from": "human", "value": "\{image\_tokens\}\\
    You are provided with some images and a question for these images. Please review the corresponding responses based on the following 5 factors:\\

    1. Accuracy in Object Description: Evaluate the accuracy of the descriptions concerning the objects mentioned in the ground truth answer. Responses should minimize the mention of objects not present in the ground truth answer, and inaccuracies in the description of existing objects.\\

    2. Accuracy in Depicting Relationships: Consider how accurately the relationships between objects are described compared to the ground truth answer. Rank higher the responses that least misrepresent these relationships.\\

    3. Accuracy in Describing Attributes: Assess the accuracy in the depiction of objects' attributes compared to the ground truth answer. Responses should avoid inaccuracies in describing the characteristics of the objects present.\\

    4. Helpfulness: Consider whether the generated text provides valuable insights, additional context, or relevant information that contributes positively to the user's comprehension of the image. Assess whether the language model accurately follows any specific instructions or guidelines provided in the prompt. Evaluate the overall contribution of the response to the user experience.\\

    \textbf{IMPORTANT INSTRUCTION:} You MUST choose either Response 1 or Response 2 as better, even if the difference is extremely subtle. "Equally good" is NOT a valid answer. If you perceive the responses as very similar in quality, you must still identify and focus on even the smallest advantages one has over the other to make your decision.\\

    You need to choose which response is better for the given question and provide a detailed reason.\\

    \textbf{Your task is provided as follows:}\\
    Question: \{sample['question']\}\\
    Answer: \{sample['answer']\}\\
    Response 1: \{modified\_steps\}\\
    Response 2: \{raw\_steps\}\\

    ASSISTANT:\\
    """
    \end{tcolorbox}
    \caption{Prompt Template for Reward Model Evaluation – llava-critic.}
    \label{prompt-llava}
\end{figure*}

\begin{figure*}[htbp]
    \centering
    \begin{tcolorbox}[colframe=red!8!white, colback=yellow!15!white, coltitle=black, title=Prompt Template for Reward Model Evaluation – R1-reward]
    \textbf{Prompt:}\\

    \texttt{"You are a highly skilled and impartial evaluator tasked with comparing two responses generated by a Large Multimodal Model for a given question.}\\

    - Start with a thorough, side-by-side comparative analysis enclosed within <think> and </think> tags. A tie is not permitted; you must choose a better option.\\

    - Conclude with a single numeric choice enclosed within <answer> and </answer> tags:\\
    \quad - Output "1" if Response 1 is better.\\
    \quad - Output "2" if Response 2 is better.\\

    \textbf{Input:}\\
    \textbf{[Question]:}\\
    \{question\_and\_answer\}\\

    \textbf{[Response 1]:}\\
    \{path1\}\\

    \textbf{[Response 2]:}\\
    \{path2\}\\

    \textbf{Output Format (strictly follow):}\\
    \texttt{<think>Your detailed comparative analysis goes here</think><answer>1/2</answer>"}
    \end{tcolorbox}
    \caption{Prompt Template for Reward Model Evaluation – R1-reward.}
    \label{prompt-r1}
\end{figure*}


    